\documentclass[11pt,a4paper]{article}
\usepackage[hyperref]{acl2020}
\usepackage{times}
\usepackage{latexsym}

\usepackage{microtype}
\usepackage{graphicx}

\aclfinalcopy 

\title{Wiki to Automotive: Understanding the Distribution Shift and its 
impact on Named Entity Recognition}

\author{Anmol Nayak, Hari Prasad Timmapathini\\
  ARiSE Labs at Bosch\\
  \texttt{\{Anmol.Nayak, HariPrasad.Timmapathini\}@in.bosch.com} \\}

\date{}

\begin{document}
\maketitle
\begin{abstract}
While transfer learning has become a ubiquitous technique used across Natural Language Processing (NLP) tasks, it is often unable to replicate the performance of pre-trained models on text of niche domains like Automotive. In this paper we aim to understand the main characteristics of the distribution shift with automotive domain text (describing technical functionalities such as Cruise Control) and attempt to explain the potential reasons for the gap in performance. We focus on performing the Named Entity Recognition (NER) task as it requires strong lexical, syntactic and semantic understanding by the model. Our experiments with 2 different encoders, namely BERT-Base-Uncased \cite{devlin-etal-2019-bert} and SciBERT-Base-Scivocab-Uncased \cite{Beltagy2019SciBERT} have lead to interesting findings that showed: 1) The performance of SciBERT is better than BERT when used for automotive domain, 2) Fine-tuning the language models with automotive domain text did not make significant improvements to the NER performance, 3) The distribution shift is challenging as it is characterized by lack of repeating contexts, sparseness of entities, large number of Out-Of-Vocabulary (OOV) words and class overlap due to domain specific nuances.
\end{abstract}

\begin{table*}
\centering
\begin{tabular}{ll}
\hline \textbf{Element} & \textbf{Details}\\ \hline
Dataset & 979 Automotive domain sentences\\
Train:Test split & 70:30\\
Random Seeds & 5\\
GPU & Nvidia Tesla V100-SXM2 (32 GB Memory)\\
NLP Library & Huggingface Transformers \cite{wolf2020transformers}\\
t-SNE & Perplexity=30, Iterations=1000\\
Optimizer for FT1 and FT3 & Adam (lr=1e-3, $\beta_1$=0.9, $\beta_2$=0.999, $\epsilon$=1e-7) \cite{kingma2014adam}\\
Optimizer for FT2 and FT4 & Adam (lr=1e-5, $\beta_1$=0.9, $\beta_2$=0.999, $\epsilon$=1e-7)\\
Optimizer for LM fine-tuning & Adam (lr=5e-5, $\beta_1$=0.9, $\beta_2$=0.999, $\epsilon$=1e-8)\\
\hline
\end{tabular}
\caption{\label{expsetup} Experiment Setup.}
\end{table*}

\section{Introduction}

Recent advancements in pre-trained networks such as BERT \cite{devlin-etal-2019-bert}, RoBERTa \cite{liu2019roberta}, GPT \cite{radford2018improving}, ERNIE \cite{sun2020ernie}, Electra \cite{clark2020electra}, T5 \cite{raffel2019exploring} etc. have lead to state-of-the-art (SOTA) performance on several NLP tasks of the GLUE Benchmark \cite{wang2018glue}. This has enabled the use of such networks for various downstream tasks with minimal task specific fine-tuning. In our work, we experiment with two different variants of BERT due to its use in scientific domain SOTA models as well. BERT has been a breakthrough in language understanding by leveraging the multi-head self-attention mechanism \cite{vaswani2017attention} in its architecture. With the Masked Language Modelling (MLM) method, it has been successful at leveraging bi-directionality while training the language model (LM), and was trained on Wikipedia (Wiki) articles. BERT-Base models have 12 encoder layers, with each layer consisting of 12 self-attention heads.

NER is the task of identifying and classifying named entities of a domain text. It relies on using lexical, syntactic and semantic information to understand and classify entities. The study of NER in the automotive domain has been very limited and mostly restricted to identifying simple entity classes like \textit{LOCATION, ORGANIZATION} etc. in the context of automotive domain \cite{rubens2002information,keraghel2020data,sivaraman2021hybrid} and very few have focused on dealing with technical description documents \cite{nayak2020knowledgekg,nayak2020knowledge,kesri2021autokg}. Further, these works have not studied recent SOTA encoders and the results with transfer learning. Task specific data is often limited and difficult to generate in niche domains like automotive as it requires a domain experts understanding to perform annotation. This has lead to using pre-trained networks for fine-tuning with minimal data, however this often is not able to replicate the performance of such pre-trained networks in domain specific settings. For example, while SciBERT \cite{Beltagy2019SciBERT}, BioBERT \cite{lee2020biobert}, PubMedBERT \cite{pubmedbert} have been able to show improvements over BERT on the NER task with scientific datasets such as SciERC \cite{luan2018multi}, NCBI-disease \cite{dougan2014ncbi}, their performance however still lags behind in comparison to the performance that BERT achieved on general domain datasets like CoNLL-2003 \cite{sang2003introduction}, ACE-2004 \cite{strassel2003multilingual}.

Hence, we wanted to understand the reasons behind this gap in performance by taking a niche domain like automotive due to our background and domain knowledge. We attempt to answer the following Research Questions for NER in the automotive domain:

\begin{enumerate}
    \item Is using a scientific domain encoder (SciBERT) preferable over a Wiki domain encoder (BERT)?
    \item Is there significant improvement in performance when the encoder's language model is fine-tuned with automotive domain text?
    \item Is fine-tuning only the head sufficient to achieve reasonable performance?
    \item What are the likely factors which prevent the performance of the model to reach 90's even under different fine-tuning settings?
\end{enumerate}

\begin{table*}
\centering
\begin{tabular}{llllll}
\hline \textbf{Class} & \textbf{IV Count}& \textbf{OOV Count}& \textbf{OOV\%}& \textbf{IV Avg. Freq}& \textbf{OOV Avg. Freq}\\ \hline
Other & 11094/11785& 1259/568 & 10.1/4.59 & 15.34/15.79 & 3.83/1.86     \\
Signal & 2899/3066& 1483/1316& 33.8/30.03 &10.2/10.22& 2.6/2.37    \\
Value & 1556/1623& 283/216& 15.38/11.74 &6.27/5.94& 1.74/1.57    \\
Action & 1244/1306& 354/292& 22.15/18.27 &4.54/4.66& 2.62/2.26    \\
Function & 943/1140& 577/380& 37.96/25 &8.89/9.82& 3.22/2.24    \\
Calibration & 350/374& 398/374& 53.2/50 &7.14/6.67& 1.8/1.75    \\
Component & 469/515& 153/107& 24.59/17.2 &6.51/6.51& 2.42/1.91    \\
State & 326/338& 212/200& 39.4/37.17 &13.58/12.51&2.65/2.59     \\
Math & 331/344& 34/21& 9.31/5.75 &6.24/5.83&2.12/2.1    \\
\hline
\end{tabular}
\caption{\label{edaner} In-Vocabulary (IV) and Out-Of-Vocabulary (OOV) analysis of the NER classes with BERT/SciBERT. Count signifies the number of times the entities of different classes were IV and OOV. Average Frequency signifies the average number of times each unique entity repeated itself.}
\end{table*}

\begin{table*}
\centering
\begin{tabular}{llll}
\hline \textbf{Encoder} & \textbf{Train Samples} & \textbf{Test Samples} & \textbf{Test Perplexity}\\ \hline
BERT & 685 & 294 & 8.748\\
    SciBERT & 685 & 294 & 7.904\\
\hline
\end{tabular}
\caption{\label{lm}Language Model Perplexity on Test dataset with BERT and SciBERT after fine-tuning with automotive domain text.}
\end{table*}

\section{Experiment Setup}

The dataset consisted of 979 annotated sentences of the automotive domain from technical documents describing functionalities like cruise control, exhaust system, braking etc. The annotation was done by automotive domain experts for 9 automotive specific NER classes: Other (words outside named entities e.g. \textit{the, in}), Signal (variables holding quantities e.g. \textit{torque}), Value (quantities assigned to signals e.g. \textit{true, false}), Action (task performed e.g. \textit{activation, maneuvering}), Function (domain specific feature e.g. \textit{cruise control}), Calibration (user defined setting e.g. \textit{number of gears}), Component (physical part e.g. \textit{ignition button}), State (system state e.g. \textit{cruising state of cruise control}) and Math (mathematical or logical operation e.g. \textit{addition}). Table~\ref{expsetup} summarizes the details. We fine-tuned the models under the following settings:

\begin{itemize}
    \item FT1: MLM is not performed on the pre-trained encoder (BERT/SciBERT). This encoder is then frozen and the token classification head is kept unfrozen for NER fine-tuning.
    \item FT2: MLM is not performed on the pre-trained encoder. This encoder and the head are kept unfrozen for NER fine-tuning.
    \item FT3: MLM is performed on pre-trained encoder with automotive domain text. This encoder is then frozen and the head is kept unfrozen for NER fine-tuning.
    \item FT4: MLM is performed on pre-trained encoder with automotive domain text. This encoder and the head are kept unfrozen for NER fine-tuning.
\end{itemize}

\begin{table*}
\centering
\begin{tabular}{llll}
\hline \textbf{Class} & \textbf{Precision} & \textbf{Recall} & \textbf{F-1}\\ \hline
Other & 83.5/91.4/81.8/91.1& 87.3/90.6/87.5/91.9 & 85.3/91/84.5/91.5     \\
Signal & 66.2/84.2/67.2/85.4& 78/89.1/78.8/89.2 & 71.5/86.6/72.3/87.2      \\
Value & 67.3/66/67.1/67.6& 40.7/60.1/40.6/58.4 & 50.6/62.7/50.5/62.6  \\
Action & 70.6/73.8/72.7/79.3& 77.1/84.4/73.6/81.2 & 73.7/78.7/73.1/80.2  \\
Function & 68.8/77.2/70.5/79.2& 63.2/74.5/57.1/77.9 & 65.8/75.6/62.9/78.5  \\
Calibration & 69.5/88/68.7/87.7& 46/82.2/48.3/82.3 & 54.3/84.8/55.4/84.6  \\
Component & 69.6/71/70.9/72.9& 54.4/62.2/53.1/69.1 & 60.9/64.8/60.4/70.4  \\
State & 67.8/82.7/66.1/80.5& 61.2/84.9/60.1/86.5 & 62.9/83.7/62.4/83.3  \\
Math & 81/92.2/81.3/92.5& 51.5/43.8/52.4/55.4 & 62.5/55.8/63.3/68.3  \\
\textbf{Macro Avg.} & \textbf{71.6/80.7/71.8/81.8}& \textbf{62.2/74.7/61.3/76.9} & \textbf{65.3/76/65/78.5} \\
\hline
\end{tabular}
\caption{\label{bertner} NER Performance with BERT under the different fine-tuning settings (FT1/FT2/FT3/FT4) averaged over 5 random seeds.}
\end{table*}

\begin{table*}
\centering
\begin{tabular}{llll}
\hline \textbf{Class} & \textbf{Precision} & \textbf{Recall} & \textbf{F-1}\\ \hline
Other & 86.5/92/87.1/92.9& 89.1/92.9/89.8/92.4 & 87.7/92.4/88.4/92.6    \\
Signal & 74.5/86.9/71.5/87.1& 79.9/89.2/83.1/89.8 & 77.1/88/76.9/88.4     \\
Value & 72/75.4/75.3/70.8& 48.9/67.2/49.2/69 & 58.2/71/59.4/69.8  \\
Action & 73.8/79.1/76.5/77.2& 80.3/83.6/80.1/83.7 & 76.9/81.3/78.2/80.3 \\
Function & 71.7/83.5/76.2/82.7& 72.4/80.1/68/82.2 & 72/81.7/71.8/82.4 \\
Calibration & 71/88.8/72.9/92.1& 60.3/88.4/53.1/86.1 & 64.9/88.5/60.6/88.9  \\
Component & 73.3/77.3/74.9/80.2& 65.4/77.5/65.5/75.6 & 69.1/77.3/69.8/77.5  \\
State & 74.1/87.1/75.1/84.7& 76/84.2/70.7/86.4 & 74.5/85.4/72.1/85.5  \\
Math & 83.4/90/85.5/90& 63.8/63.7/67.4/65.2 & 72.1/74.4/75.3/75.3 \\
\textbf{Macro Avg.} & \textbf{75.6/84.4/77.2/84.2}& \textbf{70.7/80.8/69.7/81.1} & \textbf{72.5/82.2/72.5/82.3} \\
\hline
\end{tabular}
\caption{\label{sbner} NER Performance with SciBERT under the different fine-tuning settings (FT1/FT2/FT3/FT4) averaged over 5 random seeds.}
\end{table*}

\section{Understanding the Distribution Shift}

\paragraph{Entity and Context characteristics:}As BERT variants rely on the WordPiece tokenization algorithm \cite{schuster2012japanese}, understanding the distribution of entities that are In-Vocabulary (IV) vs Out-Of-Vocabulary (OOV) is important as the meaning of a word will then be determined by how accurate are its sub-word tokens. It can be seen in Table~\ref{edaner} that OOV Count for the different entity classes varies between 9\% and 53\% for BERT and between 4\% and 50\% for SciBERT. BERT had a larger number of OOV words across all the classes in comparison to SciBERT, which we believe has been one of the contributing factors to its weaker performance, as it has previously been shown that BERT faces tokenization challenges in domain adaptation settings which leads to semantic meaning deterioration \cite{nayak2020domain,timmapathini2021probing}. Further, we can see in Table~\ref{edaner} that the average repetitions of both IV and OOV words is extremely small. This makes it challenging for the model to learn meaningful representations of words and sub-words (in the case of OOV) as both the word and its context is continuously changing. This can be attributed to the fact that technical documents and manuals are often described in a concise and factual manner, where repetition of information is discouraged.

\paragraph{Performance with different fine-tuning settings:} It can be seen in Table~\ref{bertner} and Table~\ref{sbner}, the performance gains of using a MLM fine-tuned encoder vs a vanilla encoder is negligible both in the case FT1 vs FT3 and FT2 vs FT4. Table~\ref{lm} shows the LM perplexity of the fine-tuned encoders. We believe that fine-tuning the language model with automotive domain text did not make much difference to the NER performance due to the aforementioned non-repeating phenomena of entities and context. FT2 and FT4 perform significantly superior to FT1 and FT3 as we believe that the head alone is insufficient to encapsulate the domain specific lexical, syntactic and semantic nuances over and above what the pre-trained encoder has learnt.

\begin{figure}[h]
\centering
\includegraphics[scale = 0.43]{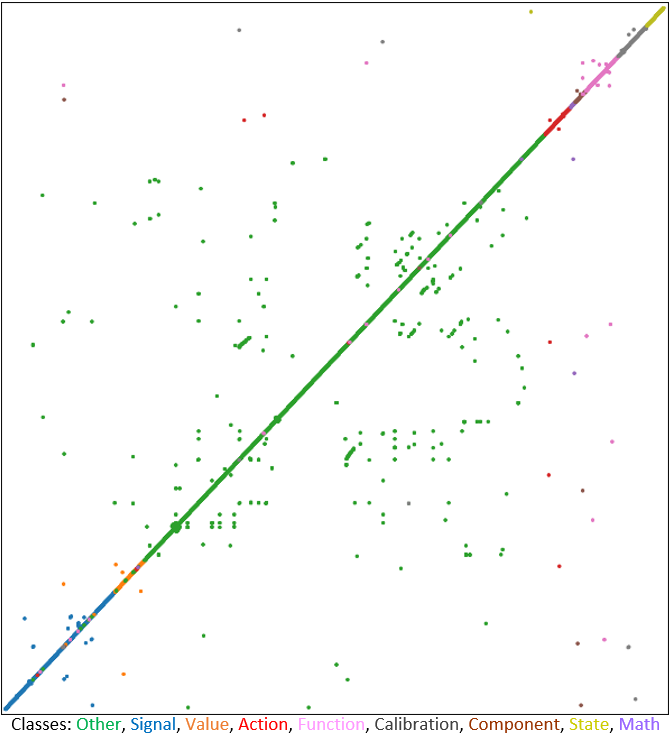}
\caption{t-SNE plot of the named entities with SciBERT embeddings (trained using FT4 setting and Random Seed 5) to understand the class overlap at last encoder layer (12).}
\label{fig:tsneplot}
\end{figure}

\paragraph{Class overlap due to domain specific notations:}

BERT has been shown to learn lexical level features in the
early layers, syntactic features in the middle layers and semantic features in the higher layers \cite{jawahar2019does}. We used the 12th layer embeddings of all the entities across the dataset and reduced them from 768 to 2 dimensions using the t-SNE algorithm \cite{van2008visualizing} for visualization (Figure \ref{fig:tsneplot}). We found that while the clusters of the classes Action, Function, Calibration, Component, State and Math remain pure (top right corner), the clusters of Other, Signal and Value begin to see overlap from other entity classes (bottom left corner).

We believe that this overlap reflected at the 12th layer is leading to a drop in performance and this is being caused at a lexical level due to use of domain specific prefixes/suffixes used across different entity classes (e.g. CrCtl can be used as a prefix across signal and state entities belonging to the cruise control function), at a syntactic level due to interchangeable grammatical usage (e.g. \textit{control} as an Action could be a verb but as a Signal could be a noun) and at a semantic level due to the use of accrued domain knowledge by the expert to make a subtle judgement that is not straightforward to infer solely from the data.

\section{Conclusion and Future Work}

In this paper we have attempted to analyze the characteristics of the distribution shift when dealing with automotive domain text by performing the NER task. We find that while fine-tuning SciBERT performs better than BERT for the automotive domain, the performance metrics are unable to touch 90's mainly due to lack of repeating contexts in the text, sparseness of entities, large number of OOV words and overlap between classes due to domain specific nuances. Our future work will focus on isolating specific attention heads and layer embeddings by passing them directly to the NER head, which previously have been shown to play unique roles in the network \cite{clark2019does}.

\bibliography{anthology,acl2020}
\bibliographystyle{acl_natbib}

\end{document}